\documentclass{ecai}  

\usepackage{graphicx}
\usepackage{latexsym}
\usepackage{bm}
\usepackage{xcolor}
\usepackage{bbm}
\usepackage{graphicx}

\begin{document}

\begin{frontmatter}

\title{Inferring Preferences from Demonstrations in Multi-Objective Residential Energy Management}

\author[A]{\fnms{Junlin}~\snm{Lu}
\thanks{Corresponding Author. Email: J.Lu5@nuigalway.ie}}
\author[A]{\fnms{Patrick}~\snm{Mannion}
}
\author[A]{\fnms{Karl}~\snm{Mason}
} 

\address[A]{University of Galway, Ireland}

\begin{abstract}


It is often challenging for a user to articulate their preferences accurately in multi-objective decision-making problems. Demonstration-based preference inference (DemoPI) is a promising approach to mitigate this problem. Understanding the behaviours and values of energy customers is an example of a scenario where preference inference can be used to gain insights into the values of energy customers with multiple objectives, e.g. cost and comfort. In this work, we applied the state-of-art DemoPI method, i.e., the dynamic weight-based preference inference (DWPI) algorithm in a multi-objective residential energy consumption setting to infer preferences from energy consumption demonstrations by simulated users following a rule-based approach. According to our experimental results, the DWPI model achieves accurate demonstration-based preference inferring in three scenarios. These advancements enhance the usability and effectiveness of multi-objective reinforcement learning (MORL) in energy management, enabling more intuitive and user-friendly preference specifications, and opening the door for DWPI to be applied in real-world settings. 
\end{abstract}

\end{frontmatter}

\section{Introduction}
The multi-objective reinforcement learning (MORL) paradigm has gained significant recognition for its effectiveness in tackling multi-objective control tasks, including energy management applications \cite{lu2022multi,mannion2018reward,zheng2022reinforcement}. For successful MORL applications, such as energy management, a user is often required to specify their preferences over objectives in the form of numerical weights. These weights are then input into a utility function to calculate the utility of various multi-objective solutions. However, accurately determining these weights can be a daunting task for users, as it necessitates a trial-and-error approach unless the user has a deep understanding of the trade-offs and inherent complexities within the domain. Even minor deviations from the true preferences can lead to unintended system behavior and outcomes that would be suboptimal for the user.

Addressing this limitation is imperative to enhance the usability and effectiveness of MORL in a wide array of domains, such as energy management. Approaches that can infer user preferences from alternative sources, such as demonstrations \cite{ikenaga2018inverse,lu2023inferring,takayama2022multi} or feedback \cite{benabbou2020interactive,benabbou2015incremental,zintgraf2018ordered}, hold great promise in making the preference specification process more intuitive and user-friendly.

In this work, we adopt the paradigm of demonstration-based preference inference (DemoPI) to infer the user's preference for energy consumption in a multi-objective context. We use the-state-of-art dynamic weight-based preference inference (DWPI) algorithm proposed by Lu et al. as the inference algorithm \cite{lu2023inferring}.

In this paper, we conduct the first study on applying the DWPI algorithm to infer preferences from real-world data. As far as we are aware, this is also the first work on inferring user preferences in the energy management domain.

\section{Background}
\subsection{Multi-Objective Reinforcement Learning}
MORL is a collection of reinforcement learning (RL) methods that are used to learn policies for multi-objective decision-making problems \cite{hayes2022practical}. In single-agent settings, MORL problems are formalized as multi-objective Markov decision processes (MOMDPs). The key difference between MOMDPs and traditional single-objective Markov decision processes (MDPs) is in how the rewards are handled.

Unlike in MDPs, the reward function in a MOMDP returns a vector to capture the rewards across multiple objectives simultaneously. To facilitate training when using single-policy learning methods, a common practice in MORL is to employ reward scalarization. This involves using a scalarization function, also known as a utility function, to map reward vectors to a scalar value. One simple and widely used example of a utility function is a linear weighted sum \cite{abels2019dynamic,barrett2008learning,castelletti2012tree,lu2022multi,lu2023inferring,roijers2017interactive}, where the weights denote the importance of each objective. With different weight vectors, different optimal policies can be learned. The set of optimal policies, i.e., the non-dominated set is known as Pareto optimal set (POS). The POS represents the best possible set of trade-off solutions, where no single solution is considered to be optimal for all objectives.

\subsection{Preference Inference}
Preference inference (PI), also known as preference learning or preference elicitation, is the process of automatically deriving users' preferences from their feedback or demonstrations, instead of explicitly requiring them to provide numerical weights.
There are two main branches of PI, i.e. inferring from feedback \cite{benabbou2020interactive,benabbou2015incremental,shao2023eliciting,zintgraf2018ordered} and inferring from demonstrations (DemoPI) \cite{ikenaga2018inverse,lu2023preference,lu2023inferring,takayama2022multi}. 

Inferring from feedback requires regular interaction with the user, which can be exceptionally time-consuming and even impractical when dealing with complex problems. We limit the scope of this work to inferring preferences from demonstrations only.

DemoPI methods like the project method (PM) proposed by Ikenaga et al. \cite{ikenaga2018inverse} and multiplicative weights apprenticeship learning (MWAL) proposed by Takayama et al. \cite{takayama2022multi} use heuristic search and a ``first train then compare" (FTTC) approach. Both methods compare the performance of a reinforcement learning (RL) agent by training it from scratch with the inferred preference to the demonstrated performance. The "performance" here is the representation of episodic reward. If the inferred preference aligns closely with the demonstrated performance, it is considered a correct inference. However, this approach has certain limitations. It necessitates training an RL agent with the inferred preference for each inference round, which can be highly time-consuming \cite{lu2023inferring}. Furthermore, the FTTC paradigm heavily assumes that the demonstration is always based on an optimal policy, which is not a realistic assumption since obtaining optimal demonstrations itself is challenging. In most cases, the inference model only receives sub-optimal demonstrations because even the user themselves have no idea about what is an optimal demonstration. Although these demonstrations still contain the necessary information to estimate a preference vector, due to the apparent difference from the agent-generated reward, the FTTC paradigm struggles to provide accurate inferences. To address these drawbacks, Lu et al. proposed the Dynamic Weight Preference Inference (DWPI) algorithm \cite{lu2023inferring}. The DWPI algorithm overcomes these limitations by employing a dynamic weight MORL (DWMORL) agent \cite{kallstrom2019tunable}, significantly improving time efficiency and achieving accurate inferences from sub-optimal demonstrations. In this study, we leverage the DWPI algorithm to infer preferences from demonstrations in the context of energy consumption because of its time efficiency and robustness.

\section{Residential Energy Consumption Model}

\subsection{Problem Description}
We use the DWPI algorithm to infer the preference over the two conflicting objectives, i.e. maximizing the comfort and saving the cost, from a demonstration in a residential energy consumption scenario. 
Two datasets are used in this residential energy simulator to provide practical environment.

\textbf{Electricity Price:} 
The electricity prices are from the PJM dataset \cite{2021PJMDataset}. The training data spans the whole day from 01/05/2021 00:00 to 02/05/2021 00:00. While the evaluation of the simulated comparison experiment uses data from 01/05/2021 00:00 to 07/05/2021 00:00, which is weekly data.

\textbf{Background Load and Renewable Generation:} 

The renewable generation and background load data are from the "Home C" of the Smart* dataset for sustainability within the UMass dataset \cite{barker2012smart}. The training data spans the whole day from 01/05/2014 00:00 to 02/05/2014 00:00. While data from 01/05/2014 00:00 to 07/05/2014 00:00 are used in the simulated comparison experiment. 

\subsection{Residential Energy Consumption Preference Inference Model}

\begin{figure}
\centering
\includegraphics[width=0.4\textwidth]{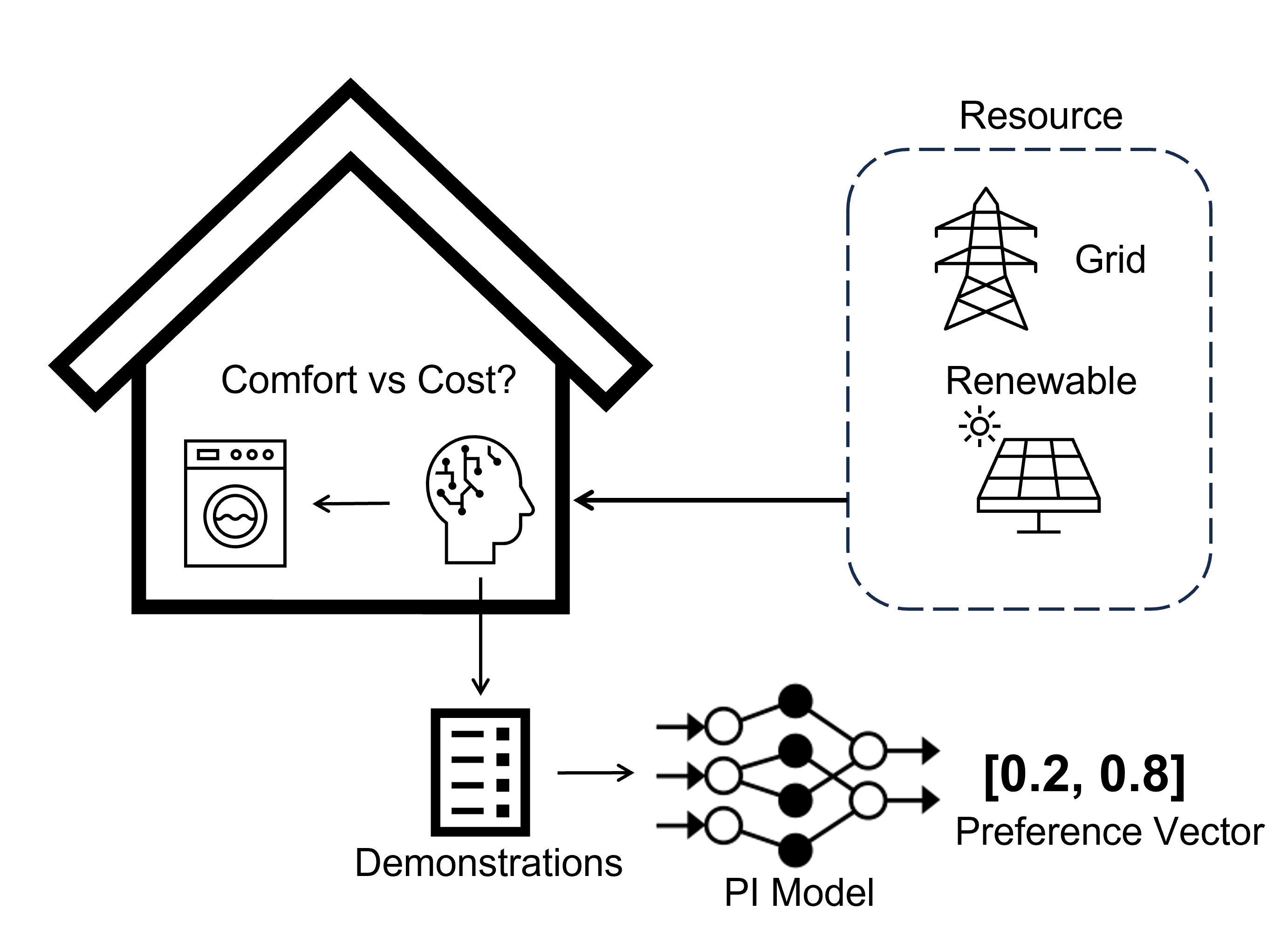}
\caption{Residential Energy Consumption Preference Inference Model. The vector [0.2,0.8] indicates that the user puts 20\% importance on maximizing comfort and 80\% importance on saving cost} \label{fig1}
\end{figure}

An overview of the PI model for energy consumption is illustrated in Figure \ref{fig1}. There are two sources of energy, i.e., the grid and the renewable generation. An agent will decide based on the state of the environment whether to start the washing machine while balancing the two objectives, maximizing comfort and saving cost. There is also background energy consumption of the refrigerator, lights, and the alarm system, etc. which have to work in a fixed routine and not influenced by the agent.

We have chosen the "Bosch WAJ28008GB Washing Machine" as the appliance being controlled by the agent. This washing machine has a power rating of 1 kW. Our assumption is that the user needs to wash clothes every day and requires them to be ready in the morning. This setup creates a dilemma for the user, requiring a trade-off between maximizing comfort and cost savings. Opting for the ideal time slot for operating the washing machine, during which it can maximize user comfort, may lead to higher costs due to the lack of renewable energy generation at night. On the other hand, running the machine during mid-day can save costs but might compromise user comfort as they cannot get clean clothes in the morning.

The user can generate some demonstrations from their underlying preference by interacting with the environment. These demonstrations are then passed to the preference inference model to infer the normalized preference weight vector. This can also be deemed as a process of quantifying their underlying thoughts over energy consumption.

\subsection{Residential Energy Consumption MOMDP}
In this section, we formalize and give the detail of the environment where the DWMORL agent and DWPI model are trained.
\subsubsection{State Space}
The state space comprises a tuple $(price,p^{r},p^{b},task^{r},t)$.
The $price$ is the average grid electricity price of the last hour, $p^{r}$ is the last hour's average renewable generation power, $p^{b}$ is the average background appliance loads in the past hour, $task^{r}$ is the hours left to be done, and $t$ is the current time of the day.

\subsubsection{Action Space}
The action set has two actions: 0 represents not running the appliance, while 1 represents running it.

\subsubsection{Reward Function}
Due to the multi-objective nature, two reward functions are introduced in this section.\\ 
The first reward is the \textbf{cost saving reward}:
\begin{equation}
    r^{cost}_{t} = -10\cdot price_{t}\cdot max[p^{s}_{t}+p^{b}_{t}-p^{r}_{t},0]
\end{equation}
where the electricity price at time t is denoted as $price_{t}$, while $p^{s}_{t}$ represents the power consumption of the shiftable loads controlled by the user or agent, i.e., the washing machine. Additionally, $p^{b}_{t}$ and $p^{r}_{t}$ correspond to the power consumed by background loads and renewable energy generation, respectively.

To calculate the reward, the electricity price at time $t$ is multiplied by the maximum value between ($p^{s}_{t} + p_{t}^{b} - p_{t}^{r}$) and 0. This formulation aims to incentivize minimizing total power consumption cost. A higher cost results in a higher penalty, encouraging both energy efficiency and the use of renewable energy sources. The constant value of 10 is introduced to ensure the cost-saving reward and comforting reward are on a similar scale.\\
The second reward is the \textbf{maximizing comfort reward}:
\begin{equation}
    r^{comf}_{t} = task^{r}_{t}\cdot action\cdot \mathbbm{1}_{t\in[0:00,7:00]}
\end{equation}
where the number of hours remaining in the task (run washing machine) at time t is denoted as $task^{r}_{t}$, while $action$ represents whether to turn on the appliance. $\mathbbm{1}_{t\in[0:00,7:00]}$ is the indicator variable and if the time is in the interval of 0:00 to 7:00 am, it is set to 1 otherwise 0. This is to guarantee that the agent can only receive a comfort reward when in the required time slot.

The reward given to the agent at each timestep is a vector $\bm{r}_{t}=[r^{cost}_{t}, r^{comf}_{t}]$ and it needs to be scalarized by a linear utility function via the preference vector, as per Eqn. \ref{eqn:scalarised_reward}. The preference weight vector is noted as $\bm{w} = [w_{cost}, w_{comf}]$.
\begin{equation}
    \label{eqn:scalarised_reward}
    r_{t} = \bm{r}_{t}\cdot \bm{w}
\end{equation}

\subsection{DWPI Model}
\begin{figure}
\centering
\includegraphics[width=0.4\textwidth]{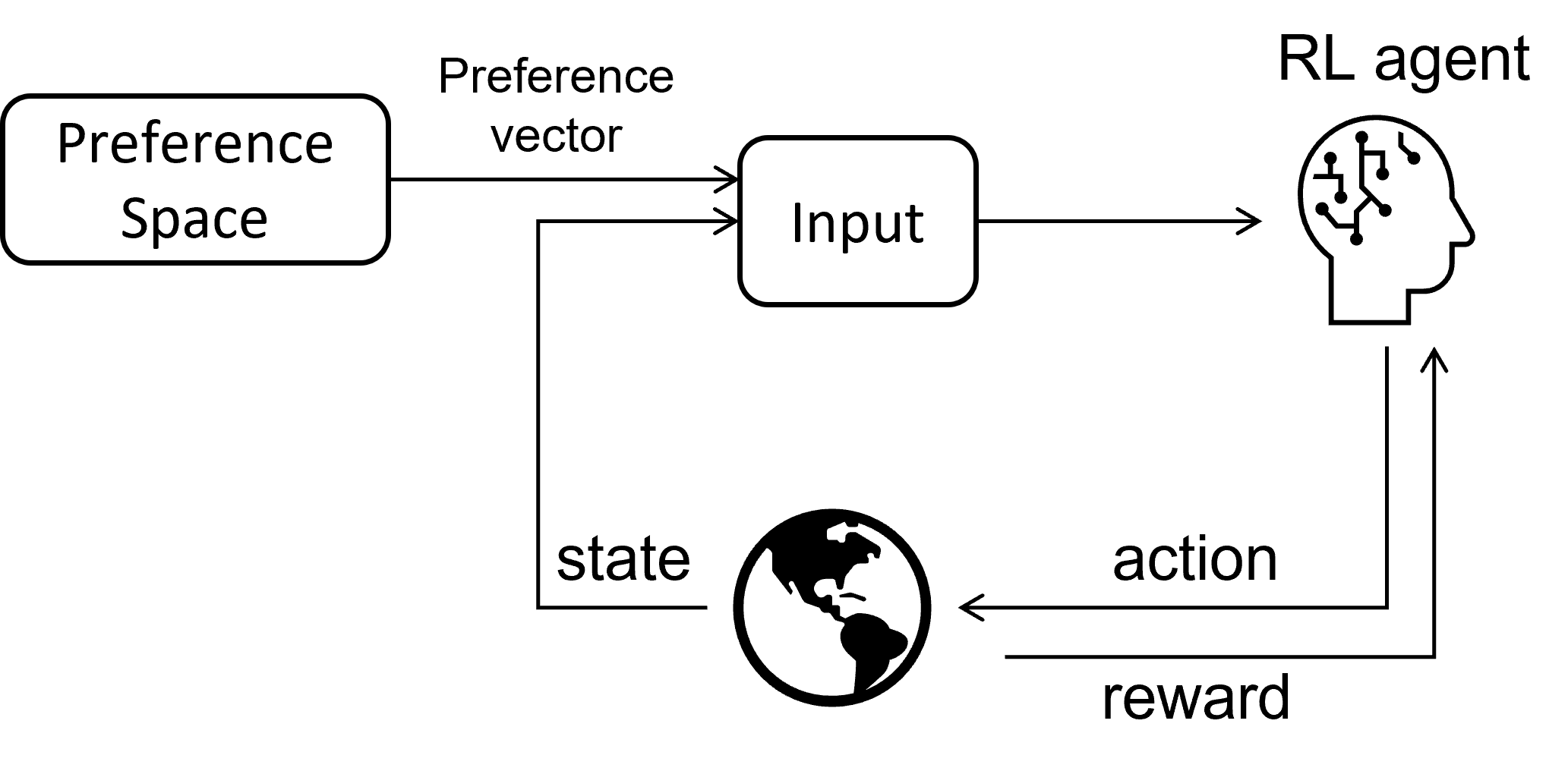}
\caption{DWMORL Agent Training} \label{DWMORL}
\end{figure}

The DWPI Model is constructed based on a DWMORL agent based on DQN that was proposed by Kallström et al. \cite{kallstrom2019tunable}. The DWMORL agent is capable of adapting its behavior based on a preference weight vector during runtime. The training process of the DWMORL agent is illustrated in Figure \ref{DWMORL}.

At the beginning of each episode, a preference weight vector is randomly selected from the preference space. This vector, along with the state obtained from the environment, is used to create an extended state input for the RL agent. The RL agent then selects an action based on this extended state and receives a reward signal from the environment. Through multiple episodes, the RL agent gradually learns to adapt its behavior based on the given preference. As a result, the agent's policy becomes dependent on the preference weight vector, enabling it to accommodate the preference during runtime without the need for retraining.

\begin{figure}
\centering
\includegraphics[width=0.4\textwidth]{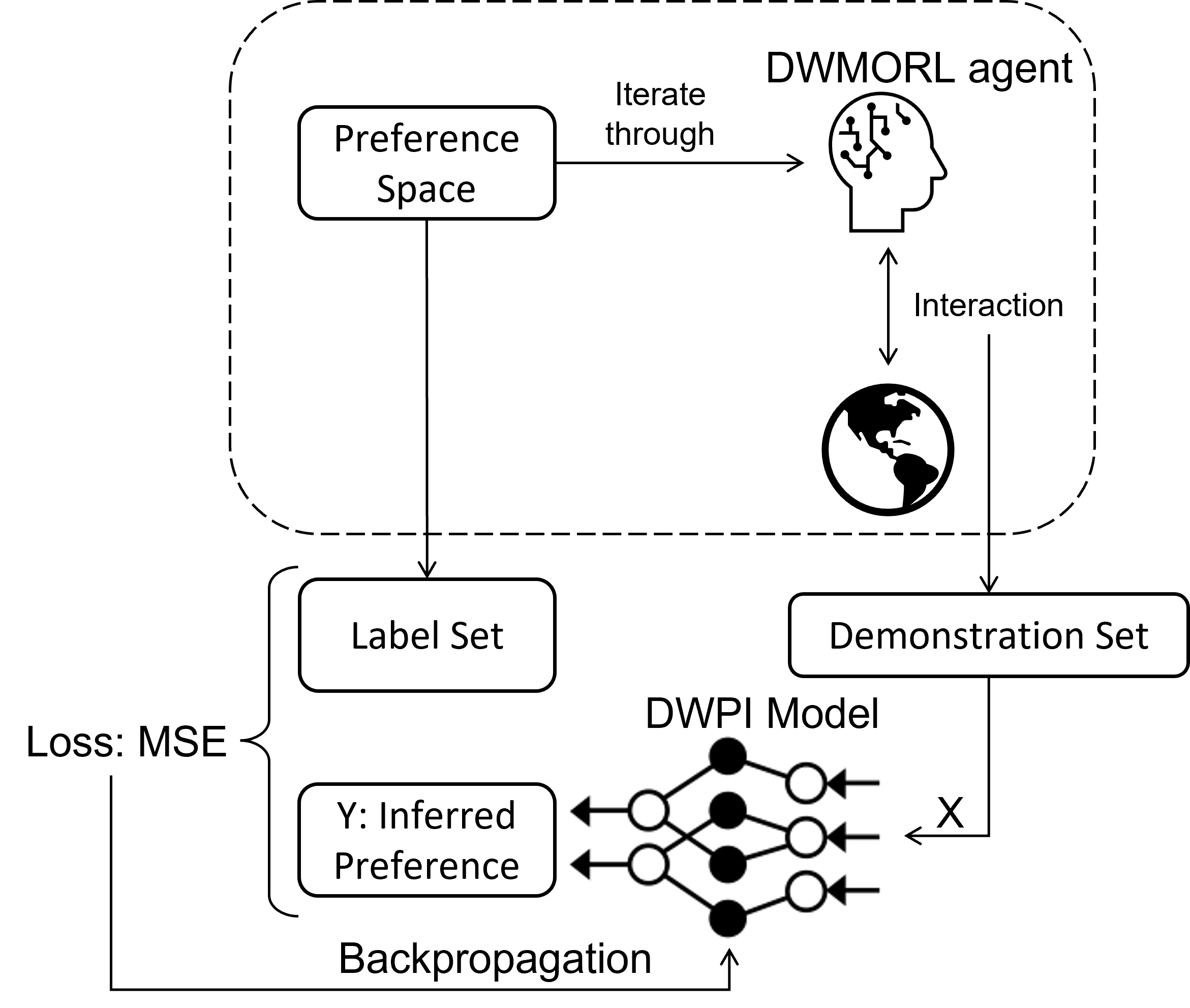}
\caption{DWPI Model Training. X is the demonstration and Y is the predicted preference vector} \label{DWPI Model Training}
\end{figure}

After the DWMORL agent is trained, we iterate through all possible preference weight vectors and generate a collection of agent demonstrations paired with the corresponding preference weights. Subsequently, the DWPI model is trained using a supervised learning approach with the demonstration representation as input and the preference weight vector as output. The training process is illustrated in Figure \ref{DWPI Model Training}. The demonstration representation, employed in this study, is the cumulative reward obtained in the trajectory. 

\section{Experiment}

\subsection{Inference Validation Experiment}
The benchmark of the DWPI agent is set to infer the user preferences in demonstrations from the three scenarios:
\begin{enumerate}
    \item \textbf{Always maximize comfort}:\\
    The demonstration for this test is issued as always running the washing machine between 2:00 am to 4:00 am, which is always within the interval of [0:00,7:00]
    \item \textbf{Always save cost}:\\
    The demonstration for this test is issued as always running the washing machine at 10:00 am to 11:00 am and 2:00 pm to 3:00 pm, which will give the lowest cost on the day. 
    \item \textbf{Mixture}:\\ 
    The demonstration for this test is issued as always running the washing machine at 6:00 am to 7:00 am (which fits the required comfort interval) and also at 10:00 am to 11:00 am (which considers both the cost and comfort). 
\end{enumerate}
As the Pareto front of this problem is unknown, we evaluate the accuracy of the preference inference method in this problem qualitatively. For the "always maximize comfort" scenario, the inferred weight for comfort should be much higher than for cost-saving. In the "always save cost" scenario, the inferred weight for cost-saving should be much higher than the weight for comfort. In the "mixture" scenario, the weight factors for both objectives should be similar.

\subsection{Simulated Comparison Experiment}
After inferring the preference from the three scenarios, the inferred preference is utilized as a fixed preference vector. This fixed preference vector is then provided as input to a MORL agent, which is trained using the same data (one-day data) as the DWMORL agent. The purpose of incorporating the fixed preference vector is to enhance the agent's policy by ensuring a stable training process. Subsequently, the trained MORL agent is deployed to simulate in a 7-day period. The agent-generated result is compared with the user-generated result. The user-generated result is on the same 7-day data, during which the user operates the appliance in alignment with the three scenarios introduced (rule-based). If the agent-generated result is close to the user-generated result, the preference is further proven to be reasonably accurate. 

\subsection{Hyperparameters}
The hyperparameters of both the DWMORL agent are shown in Table \ref{DWMORL Hyperparameters}.
\begin{table}
\centering
\caption{DWMORL Agent Hyperparameters}\label{DWMORL Hyperparameters}
\begin{tabular}{c|c}
\hline
Hyperparameter & \\
\hline
RL Algorithm & DQN\\
Number of Episodes &  20000 \\
Replay Memory Size & 1000 \\
Epsilon Decay & $1/(episode\cdot 0.98)$\\
Epsilon Start & 1\\
Epsilon End & 0.01\\
Start Training After (Episodes)& 10\\
Copy to Target Per (Episodes) & 50\\
Batch Size & 64\\
Gamma & 1\\
Learning Rate & 0.001\\
Loss Function & MSE \\
Hidden Layer Structure & [32,32,16]\\
\hline
\end{tabular}
\end{table}

The hyperparameters of both the DWPI model are shown in Table \ref{DWPI Hyperparameters}.

\begin{table}
\centering
\caption{DWPI Model Hyperparameters}\label{DWPI Hyperparameters}
\begin{tabular}{c|c}
\hline
Hyperparameter & \\
\hline
Number of Epochs &  1500 \\
Batch Size & 32\\
Gamma & 1\\
Learning Rate & 0.01\\
Loss Function & MSE \\
Hidden Layer Structure & [16,16,8]\\
\hline
\end{tabular}
\end{table}

\section{Results and Discussion}
The results of the DWPI model in the three benchmark scenarios are shown in Table \ref{tab1}.
\begin{table}
\centering
\caption{Inference Validation Experiment Result}\label{tab1}
\begin{tabular}{c|c|c}
\hline
Scenario &  Demonstration - running at & $[w_{cost}, w_{comf}]$\\
\hline
Always max comfort &  2am - 4am &[0.26, \textbf{0.74}]\\
Always save cost & 10am - 11am \& 2pm - 3pm& [\textbf{0.79}, 0.21]\\
Mixture & 6am - 7am \& 10am - 11am&[0.44, \textbf{0.56}]\\
\hline
\end{tabular}
\end{table}
The highest weight is in \textbf{bold}. In the demonstration from the "always maximize comfort" scenario, the preference is inferred as [0.26, 0.74]. Notably, the weight assigned to maximizing comfort is approximately 0.74, which is nearly three times higher than the weight for saving costs. This indicates a strong emphasis on maximizing comfort in this scenario.

Similarly, in the demonstration from the "always save cost" scenario, the preference is inferred as [0.79, 0.21]. In this case, the weight for saving cost is approximately four times higher than the weight for maximizing comfort. This suggests a significant emphasis on cost-saving in this scenario. These preference inferences clearly highlight the contrasting priorities in the two scenarios, with one scenario focusing predominantly on comfort and the other on cost-saving. 

In the final inference, which corresponds to the "mixture" scenario, the preference of the demonstration is inferred as [0.44, 0.56]. Notably, while the objective of maximizing comfort still holds greater weight, the two weight factors are now much closer to each other. This outcome is in alignment with our initial assumption for the "mixture" scenario, where a balanced emphasis on both comfort and cost-saving was anticipated. The inference result reflects the desired balance between the two objectives. From the inference results, the DWPI model is proved to be able to effectively capture these distinctions between the user demonstrations, providing valuable insights into the user's preferences for each scenario.

The outcome of the simulated comparison experiment is presented in Table 4. The results are represented as cumulative reward vectors, denoted as $[\sum r_{cost}, \sum r_{comf}]$. The units of currency are US \$. Note that currency values are negative as we wish to minimize costs. Consequently, for results on both objectives, higher values are better. 

\begin{table}
\centering
\caption{Simulated Comparison Experiment Results}\label{Simulated Comparison Experiment Result}
\begin{tabular}{c|c|c}
\hline
Scenario &  User-generated Result & Agent-generated Result\\
\hline
Always max comfort &[-22.75, 18]&[-22.22, 17]\\
Always save cost &[-22.15, 0]&[-20.40, 0]\\
Mixture &[-22.09, 12]&[-21.64, 14]\\
\hline
\end{tabular}
\end{table}

The results of the weekly data simulation indicate a close resemblance between the agent-generated and user-generated outcomes. This demonstrates that the inference model is capable of accurately inferring the user's underlying preferences and achieving similar performance as the user. 

In the scenario where the objective is to "Always maximize comfort", the agent achieves lower costs compared to the user but at the expense of reduced comfort. This occurs because the user had an extreme preference for maximizing comfort, while the agent's policy is based on the inferred preference vector [0.26, 0.74], which still allows for some consideration of cost-saving. 

Conversely, in the "Always save cost" scenario, the agent achieves greater cost savings than the user. In the "Mixture" scenario, the agent also produces results similar to the user.

Comparing the three scenarios, a clear descending trend is observed in terms of comfort maximization, ranging from "Always max comfort" to "Mixture" to "Always save cost." Conversely, there is an ascending trend in terms of cost-saving. These findings indicate that the inferred preference significantly influences the agent's behavioral patterns, aligning them with the corresponding user demonstrations.

\section{Conclusion}
This study addresses the challenge of numerically specifying preferences in MORL for energy management. By using DemoPI, the DWPI algorithm accurately captures user preferences in different scenarios. We explored the inference of user preferences and their impact on the behavior of an agent in the context of energy consumption. As shown in the original DWPI paper \cite{lu2023inferring}, DWPI is much more time efficient than other methods to infer preferences from demonstrations, making it a good candidate for preference inference in real world problems. 

We introduce three scenarios to do the preference inference, i.e., "Always maximise comfort", "Always save cost", and "Mixture". In the previous two scenarios, the user has an extreme preference for maximizing comfort and saving cost, while in the third scenario, the user would like to balance the two objectives. The result of the inference validation experiment demonstrates that the DWPI model can accurately infer the user's preference qualitatively. 

Further, in the simulated comparison experiment, the inferred preferences were then incorporated as fixed preference vectors in training a MORL agent to see if the MORL agent can achieve similar results as the user. The results show that the inference model successfully captured the underlying user preferences, achieving a similar performance to that of the user. Comparing the three scenarios, the results indicated a clear trade-off between maximizing comfort and saving costs. The preference inferred from the user's demonstrations significantly influenced the agent's behavior patterns, aligning them with the corresponding user preferences. However, slight deviations were observed due to the agent learning based on the inferred preferences, which may have differed slightly from the user's extreme preferences. 

The PI methods can be further refined, e.g., by using a more advanced DWMORL agent than the DQN-based agent in the present work to increase the inference accuracy and the model's robustness. More complex scenarios of appliance scheduling are another extending direction, where multiple objectives such as switching on and off the heating, air conditioning, and sweeping robots can be considered. Other further research can be, incorporating demonstrations from real human users rather than simulated users, and extending this work into a multi-agent context.


\ack This research is funded by an Irish Research Council Government of Ireland Postgraduate Scholarship (GOIPG/2022/2140).

\bibliography{ecai}
\end{document}